\documentclass[letterpaper,10pt,conference]{ieeeconf}

\IEEEoverridecommandlockouts %
\usepackage{times}    %
\usepackage{amsmath}  %
\usepackage{amssymb}  %
\usepackage{bm}       %
\usepackage{booktabs}
\usepackage{multirow}
\usepackage{array}
\usepackage{epsfig}
\usepackage{etoolbox}
\usepackage{graphics}
\usepackage{makecell}
\usepackage{microtype}
\usepackage{stfloats}

\usepackage{algorithm, algpseudocode}

\usepackage{mathtools}  %
\DeclarePairedDelimiter\abs{\lvert\lvert}{\rvert\rvert}%

\usepackage[noadjust]{cite} %
\usepackage{subcaption}
\usepackage[bookmarks=true,hidelinks]{hyperref}
\usepackage{paralist}  %

\usepackage[binary-units=true]{siunitx}
\usepackage{cleveref}

\newcommand{\fref}[1]{Fig.~\ref{#1}}
\newcommand{\eref}[1]{Eq.~(\ref{#1})}

\newcommand{\vq}{\bm{q}}
\newcommand{\vqdot}{\bm{v}}
\newcommand{\vqddot}{\dot{\bm{v}}}
\newcommand{\vx}{\bm{x}}

\newcommand{\vu}{\bm{u}}
\newcommand{\vv}{\bm{v}}

\newcommand{\bM}{\bm{M}}

\newcommand{\vlam}{\pmb{\lambda}}

\newcommand{\argmin}{\operatornamewithlimits{arg\ min}}

\usepackage[dvipsnames]{xcolor}

\definecolor{defaultblue}{RGB}{0, 0, 255}
\definecolor{barblue}{RGB}{92, 192, 217}
\definecolor{groupblue}{RGB}{230,230,255}
\definecolor{linkred}{RGB}{165,0,33}

\usepackage[latin1]{inputenc}

\usepackage{graphicx}
\graphicspath{{figures/}}
\usepackage[export]{adjustbox} %

\usepackage[style=long,nolist]{glossaries}
\glsdisablehyper
\newacronym{sparsefddp}{SparseFDDP}{Sparsity-inducing feasibility-driven Dynamic Differential Programming}

\newacronym{c-space}{$\mathcal{C}$-space}{configuration space}
\newacronym{com}{CoM}{Centre-of-Mass}
\newacronym{dof}{DoF}{degrees of freedom}
\newacronym{eom}{EoM}{equation of motion}
\newacronym{fk}{FK}{forward kinematics}
\newacronym{giw}{GIW}{Gravito-Inertial Wrench}
\newacronym{ik}{IK}{inverse kinematics}

\newacronym{hmm}{HMM}{Hidden Markov Model}

\newacronym{lq}{LQ}{Linear-Quadratic}
\newacronym{lp}{LP}{linear program}
\newacronym{lping}{LP}{Linear Programming}
\newacronym{nlp}{NLP}{Non-linear Programming}
\newacronym{qp}{QP}{Quadratic Programming}
\newacronym{sdp}{SDP}{Semidefinite Programming}
\newacronym{sqp}{SQP}{Sequential Quadratic Programming}
\newacronym{miqcqp}{MIQCQP}{Mixed-Integer Quadratically Constrained Quadratic Programme}
\newacronym{sip}{SIP}{Semi-Infinite Programming}

\newacronym{lbfgs}{L-BFGS}{Limited-memory BFGS}
\newacronym{bfgs}{BFGS}{Broyden-Fletcher-Goldfarb-Shanno}
\newacronym{prm}{PRM}{Probabilistic Roadmap}
\newacronym{rrt}{RRT}{Rapidly-Exploring Random Tree}
\newacronym{hdrm}{HDRM}{Hierarchical Dynamic Roadmap}
\newacronym{drm}{DRM}{Dynamic Roadmap}
\newacronym{gjk}{GJK}{Gilbert--Johnson--Keerthi}
\newacronym{epa}{EPA}{Expanding Polytope Algorithm}
\newacronym{cmaes}{CMA-ES}{Covariance Matrix Adaptation Evolution Strategy}
\newacronym{pso}{PSO}{Particle Swarm Optimization}
\newacronym{chomp}{CHOMP}{Covariant Hamiltonian Optimization for Motion Planning}
\newacronym{stomp}{STOMP}{Stochastic Trajectory Optimization for Motion Planning}
\newacronym{aico}{AICO}{Approximate Inference COntrol}
\newacronym{riemo}{RieMo}{Riemannian Motion Optimization}
\newacronym{komo}{KOMO}{k-Order Motion Optimization}
\newacronym{pi2}{$\text{PI}^2$}{Policy Improvement with Path Integrals}
\newacronym{trajopt}{TrajOpt}{Trajectory Optimization for Motion Planning}
\newacronym{ddp}{DDP}{Differential Dynamic Programming}
\newacronym{sbp}{SBP}{Sampling-based Planning}
\newacronym{cg}{CG}{Conjugate Gradient}
\newacronym{knn}{KNN}{$k$-Nearest Neighbor}
\newacronym{pca}{PCA}{Principal Component Analysis}
\newacronym{mse}{MSE}{Mean Squared Error}
\newacronym{mae}{MAE}{Mean Absolute Error}
\newacronym{lwpr}{LWPR}{Locally Weighted Projection Regression}
\newacronym{gpr}{GPR}{Gaussian Process Regression}

\newacronym{admm}{ADMM}{Alternating Direction Method of Multipliers}
\newacronym{mpc}{MPC}{Model-Predictive Control}
\newacronym{pbd}{PbD}{Programming by Demonstration}
\newacronym{lfd}{LfD}{Learning from Demonstration}
\newacronym{ioc}{IOC}{Inverse Optimal Control}
\newacronym{irl}{IRL}{Inverse Reinforcement Learning}
\newacronym{oc}{OC}{Optimal Control}
\newacronym{rl}{RL}{Reinforcement Learning}

\newacronym{moe}{MoE}{Mixture-of-Experts}
\newacronym{poe}{PoE}{Product-of-Experts}
\newacronym{gmm}{GMM}{Gaussian Mixture Model}

\newacronym{scd}{SCD}{Smooth Collision Distance}
\newacronym{sop}{SoP}{Sum-of-Penetrations}
\newacronym{cd}{CD}{Collision Distance}
\newacronym{cc}{CC}{Collision Check}

\newacronym{sme}{SME}{Small and Medium Enterprise}
\newacronym{agv}{AGV}{Autonomous Ground Vehicle}
\newacronym{ros}{ROS}{Robot Operating System}
\newacronym{imu}{IMU}{Inertial Measurement Unit}
\newacronym{gps}{GPS}{Global Positioning System}
\newacronym{slam}{SLAM}{Simultaneous Localisation and Mapping}
\newacronym{ukf}{UKF}{Unscented Kalman Filter}
\newacronym{rsi}{RSI}{Repetitive Strain Injury}
\newacronym{oem}{OEM}{Original Equipment Manufacturer}
\newacronym{drc}{DRC}{{DARPA} Robotics Challenge}
\newacronym{eod}{EOD}{Explosive Ordnance Disposal}
\newacronym{idrm}{iDRM}{inverse Dynamic Reachability Map}
\newacronym{mit}{MIT}{Massachusetts Institute of Technology}
\newacronym{exotica}{EXOTica}{Extensible Optimization Toolset}

\usepackage[english]{babel}

\usepackage{todonotes}

\usepackage{tikz}
\usepackage{textcomp}
\usepackage{hyperref}
\usepackage{lipsum}

\newcommand\copyrighttext{%
  \footnotesize \textcopyright This work has been submitted to the IEEE for possible publication. Copyright may be transferred without notice, after which this version may no longer be accessible.}
\newcommand\copyrightnotice{%
\begin{tikzpicture}[remember picture,overlay]
\node[anchor=south,yshift=10pt] at (current page.south) {\fbox{\parbox{\dimexpr\textwidth-\fboxsep-\fboxrule\relax}{\copyrighttext}}};
\end{tikzpicture}%
}

\title{\LARGE \bf
    A Versatile Co-Design Approach For Dynamic Legged Robots
}

\author{
    Traiko Dinev\,\,\, Carlos Mastalli\,\,\, Vladimir Ivan\,\,\, Steve Tonneau\,\,\, Sethu Vijayakumar
    \thanks{All authors are with the Edinburgh Centre for Robotics, University of Edinburgh, UK.}%
    \thanks{Carlos Mastalli is also with the School of Engineering and Physical Sciences, Heriot-Watt University, U.K.}
    \thanks{This research was supported by (1) the European Commission under the Horizon 2020 project Memory of Motion (MEMMO, ID: 780684) and (2) the Engineering and Physical Sciences Research Council (EPSRC), and (3) the Alan Turing Institute.}
}

\begin{document}
\bstctlcite{IEEEexample:BSTcontrol}

\maketitle
\copyrightnotice
\thispagestyle{empty}
\pagestyle{empty}

\begin{abstract}
    We present a versatile framework for the computational co-design of legged robots and dynamic maneuvers.
    Current state-of-the-art approaches are typically based on random sampling or concurrent optimization.
    We propose a novel bilevel optimization approach that exploits the derivatives of the motion planning sub-problem (i.e., the lower level).
    These motion-planning derivatives allow us to incorporate arbitrary design constraints and costs in an general-purpose nonlinear program (i.e., the upper level).
    Our approach allows for the use of any differentiable motion planner in the lower level and also allows for an upper level that captures arbitrary design constraints and costs.
    It efficiently optimizes the robot's morphology, payload distribution and actuator parameters while considering its full dynamics, joint limits and physical constraints such as friction cones.
    We demonstrate these capabilities by designing quadruped robots that jump and trot. We show that our method is able to design a more energy-efficient Solo robot for these tasks.
\end{abstract}

\section{Introduction}
To design a robot capable of executing dynamic motions, we need to consider the robot's mechanical design as well as the motion it will execute.
A traditional approach is to iterate between mechanical design and motion planning (e.g.,~\cite{semini_hyq_2011}). However, it is a challenging process, especially for complex and dynamic robots, as it requires experts both in motion planning and mechanical design.
Instead, \textit{concurrent design} (co-design~\cite{li_design_2001}) aims to automate this process by numerically optimizing both the motion and design parameters.
As the designer, we first specify a set of design parameters (e.g., morphologies or motor characteristics), constraints (e.g., collision avoidance between robot components), high-level tasks (e.g., a jump) and evaluation metrics (e.g., energy).
The algorithm then finds optimal design parameters and motions to more efficiently execute the task.

For the algorithm to find realistic design improvements, it needs to be able to plan feasible motions by considering the robot's full-body dynamics and actuation limits.
We can do it efficiently through motion planning frameworks such as \textsc{Crocoddyl}~\cite{mastalli_crocoddyl_2020}, which can run fast enough for predictive control applications~\cite{mastalli-mpc22}. 
On the other hand, from a designer standpoint, we need to be able to specify arbitrary design constraints and cost functions in order to give the designer tools to fully specify all the parameters of the design.

Re-implementing motion planning in order to add additional design parameters requires considerable technical work, which is why we seek a modular framework that exploits state-of-the-art motion planners while considering design constraints. With this motivation in mind, we developed a co-design algorithm with the following scope: 1) ability to define arbitrary costs and constraints on continuous design variables, 2) treat the motion planning as a module, and 3) exploit state of the art motion planners that can compute dynamic motion for legged robots which include constraints on the motion parameters. This scope has some subtle differences from other co-design work in the literature.

\begin{figure}[t]
    \centering
    \includegraphics[width=0.485\textwidth,trim={0.0cm 0.0cm 0.0cm 0.0cm},clip]{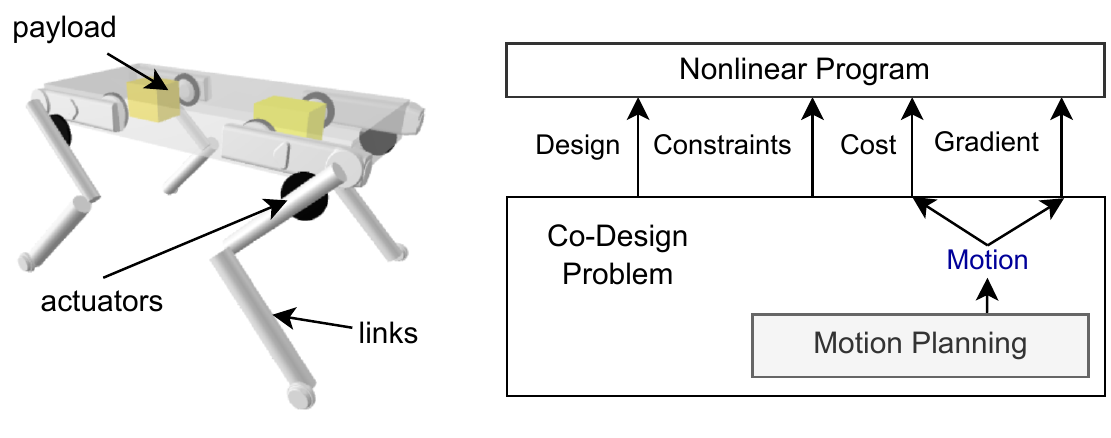}
    \caption{Illustration of our bilevel optimization approach for robot co-design. In the upper level we use gradient information from the motion planning (lower level) to optimize the design of the robot.
    Please find the accompanying video at  \url{https://youtu.be/Yxn7K1HXt_I}}
    \label{fig:poster}
\end{figure}

\subsection{Related Work}

In the current literature, a popular approach to co-design is what we call \textit{sampling-based co-design}. These methods are two-staged and exploit variants of Monte-Carlo sampling to find candidate robot designs.
The fitness of those candidates is evaluated in a second stage through a motion planner.

The Covariance Matrix Adaptation Evolutionary Strategy (CMA-ES)~\cite{wampler_optimal_2009} is a popular sampling approach used in co-design.
It uses a Gaussian prior on candidate design parameters and estimates a covariance matrix needed for the following sampling steps. For instance, Wampler et al.~\cite{wampler_optimal_2009} used a variant of CMA-ES to co-design various creatures in simulation, and Digumarti et al.~\cite{digumarti_concurrent_2014} co-designed the legs of the quadruped StarlETH to optimize its running speed. Ha et. al.~\cite{ha2016task} used CMA-ES to optimize design and swing trajectories of planar legged robots. Most recently, Chadwick et al.~\cite{chadwick_vitruvio_2020} optimized the legs of quadrupeds and bipeds over uneven terrain for different user-defined co-design metrics, and Fadini et al.~\cite{fadini_computational_2020} computed the actuator properties of a monoped using CMA-ES. 

A benefit of the above approaches is that they can use non smooth motion planners in the lower level. However, they do not support hard constraints on the design in the upper level, requiring soft constraints and cost tuning. Moreover, the algorithmic complexity of CMA-ES scales exponentially with respect to the number of design parameters (i.e., decision variables) due to the curse of dimensionality~\cite{omidvar2010comparative}, \cite{hansen2009benchmarking}.
This limits its application to a reduced number of design parameters and constraints, which in turn limits its scalability, for instance to multiple tasks and environments.

On the other hand, a number of \textit{gradient-based co-design} methods have been proposed in the literature. One approach is to formulate a single nonlinear program that optimizes both motion and design parameters. This approach has been used to co-design legged robots.
For instance, Mombaur~\cite{mombaur_using_2009}, Buondonno et al.~\cite{buondonno_actuator_2017} and Spielberg et al.~\cite{spielberg_functional_2017} compute the motions, lengths of the robot's limbs and/or actuator parameters in a single nonlinear program. However, the algorithmic complexity of the resulting nonlinear program is its major drawback (e.g. \cite{sun2020fast}).
It also requires to modify the motion planning when including new co-design requirements, making the method non-modular.

To tackle the above-mentioned drawbacks, a few recent pieces of work have proposed a new approach that uses derivative information obtained via sensitivity analysis. Ha et al.~\cite{ha_computational_2018} proposed to extract the relationship between motion and design by using the implicit function theorem.
This allowed them to optimize the design while keeping the motion on the manifold of optimal motions.
In a similar fashion, Desai et al.~\cite{desai_interactive_2018} used sensitivity analysis and the adjoint method to determine the same relationship.
This latter approach was used in~\cite{geilinger_skaterbots:_2018} and~\cite{geilinger_computational_2020} for human-in-the-loop design optimization of robots.
Still, these approaches have limitations.
For instance, the method presented in \cite{ha_computational_2018} optimizes one target design parameter at a time and requires user input to select that parameter during optimization. The approaches used in~\cite{desai_interactive_2018,geilinger_skaterbots:_2018,geilinger_computational_2020} do not impose hard constraints in the motion optimization, but rather use penalty costs. This has the potential of violating the physics constraints. Finally, none of these methods support design constraints, which is a key designer requirement.

\subsection{Our approach}
In this paper, we propose a related, but more general solution, where we directly take the derivative of the motion planner and embed it into a nonlinear program.
Our approach contains an upper and a lower level optimization for robot design and motion planning, respectively.
In the lower level, we use an efficient state-of-the-art constrained motion planner, which is continuously differentiable.
In the upper level, we formulate the design constraints and metrics as a nonlinear program, which we solve with a general-purpose nonlinear optimization software that handles arbitrary constraints.

Our approach is modular for differentiable motion planners, similar to genetic algorithms, while also supporting hard constraints on design parameters, which genetic algorithms do not. Since it uses derivative information, it inherently has faster local convergence. Finally, it does not require unconstrained motion planning (as is the case in \cite{geilinger_skaterbots:_2018}).

\subsection{Contributions}

The main contribution of our work is a novel bilevel optimization approach for robot co-design (\fref{fig:poster}). We identify two technical contributions:
\begin{enumerate}[i.]
    \item a modular co-design algorithm that differentiates a motion planner and handles arbitrary co-design constraints and metrics in the upper level;
    \item a complete co-design framework for quadruped robots and dynamic locomotion maneuvers;
\end{enumerate}

Our approach is of practical interest, as it allows for the use of any differentiable motion planner in the lower level without any modification to the motion planning itself. A modular approach like ours can take advantage of the state-of-the-art motion planning algorithms in terms of their convergence via the efficient use of the problem structure, and their ability to solve complex problems involving full robot dynamics and contacts with the environment. We show that gradient information and a bilevel optimization is a feasible approach to co-design for real-world co-design problems.

\section{Co-Design Framework}

\begin{figure}[t]
    \centering
    \includegraphics[width=0.5\textwidth, trim={0.0cm 0.0cm 0.0cm 0.0cm},clip]{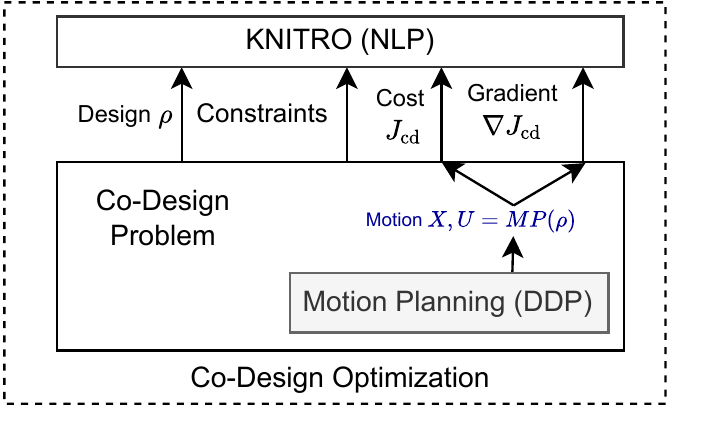}
    \caption{A schematic of our co-design framework. We optimize the robot's design by differentiating through a motion planner, which is the lower-level problem of our bi-level formulation.}
    \label{fig:pipeline}
\end{figure}

Our co-design framework is illustrated in \autoref{fig:pipeline}. First we describe our generic bilevel formulation of the co-design problem. We then describe the lower motion planning level, followed by how we apply our formulation for the design of quadrupeds. Finally, we describe a validation phase of our framework in simulation.

\subsection{Co-design as Bilevel Optimization}
We begin by encoding the robot's design into a \textit{design vector} $\bm{\rho}$. The vector $\bm{\rho}$ encodes the robot's link lengths and its base shape (width, height, depth), as well as the payload distribution and the actuator parameters -- motor mass and gear ratio.
We then formulate the co-design problem over the design vector $\bm{\rho}$ as a bilevel optimization problem:
\begin{align}
    \min_{\bm{\rho}, \bm{X}, \bm{U}} &
        \ J_\text{cd}(\bm{\rho}, \bm{X}, \bm{U}) \tag*{(co-design metric)} \\
    \text{s.t.}\ &\bm{X}, \bm{U} = \textsc{MP}(\bm{\rho}; \textsc{Task}), \tag*{(motion planning)} \\
    &  \bm{\underline{\rho}} \leq \bm{\rho} \leq {\bm{\bar{\rho}}}, \tag*{(design bounds)}\\
    & \bm{g}(\bm{\rho}) = 0, \tag*{(equality design constraints)} \\
    & \bm{h}(\bm{\rho}) \leq 0, \tag*{(inequality design constraints)} \\
    \label{eq:codesign_formulation}
\end{align}
where $J_\text{cd}(\cdot)$ is a user-specified co-design metric that evaluates the goodness of the design through the efficiency of the motion (e.g., the total energy used),
\textsc{MP}$(\cdot)$ is the motion planning function, $\bm{\underline{\rho}}$ and ${\bm{\bar{\rho}}}$ are the lower and upper bounds of the design parameters. $\bm{g}(\bm{\rho})$ and $\bm{h}(\bm{\rho})$ are general equality and inequality constraints on the design vector (e.g., no collision constraints).
We formulate the \textsc{MP} function as a nonlinear optimal control problem, which computes a discrete trajectory of robot states $\bm{X} = \{\vx_0, \dots, \vx_N\}$ and controls $\bm{U} = \{ \vu_0, \dots, \vu_{N - 1} \}$ for a desired task (represented by \textsc{Task}) such as a trotting or jumping gait. Here $N$ is the planning horizon, which is part of the task description.

We consider the motion planner as a general nonlinear function that maps from design parameters $\bm{\rho}$ to motions ${\bf{m}}=\{\bm{X}, \bm{U}\}$. Thus, we can write the derivative of the co-design cost as:
\begin{align}\nonumber
    \nabla_{\bm{\rho}} J_\text{cd} &= \frac{d J_\text{cd}(\bm{\rho}, \bf{m})}{d \bm{\rho}} = 
        \frac{\partial J_\text{cd}}{\partial \bf{m}} \frac{\partial \bf{m}}{\partial \bm{\rho}} + \frac{\partial J_\text{cd}}{\partial \bm{\rho}} \\ &=
        \frac{\partial J_\text{cd}}{\partial \bf{m}} \frac{\partial \textsc{MP}(\bm{\rho}; \textsc{Task})}{\partial \bm{\rho}} + \frac{\partial J_\text{cd}}{\partial \bm{\rho}},
    \label{eq:jmp}
\end{align}
where $\frac{\partial \bf{m}}{\partial \bm{\rho}}$ is the derivative of the motion with respect to the design parameters. This derivative can be computed using sensitivity analysis of the motion planner itself. However, the resulting expression is dependent on the optimization used in the lower level and thus not a modular solution. Computing it is also cumbersome as it  involves differentiating through a complex nonlinear program.

Instead, we can directly consider the derivative $\frac{\partial \textsc{MP}(\bm{\rho}; \textsc{Task})}{\partial \bm{\rho}}$, or even more generally, directly $\nabla_{\bm{\rho}} J_\text{cd}$. This derivative would be difficult to compute analytically, however in practice the dimension of $\bm{\rho}$ is small, compared to the dimension of the motion. For instance, in our trotting experiment, $\text{dim}(\bm{\rho}) = 17$, while $\text{dim}(\bm{m}) = 9163$.  
Hence we can obtain $\nabla_{\bm{\rho}} J_\text{cd}$ directly through a numerical differentiation procedure that runs in parallel, i.e., we compute the derivative for each component of the $\bm{\rho}$ using multiprocessing. Using a one-sided finite difference approach, this requires a total of $\text{dim}(\bm{\rho}) + 1$ calls to the motion planner. For each component of $\bm{\rho}$, we have:
\begin{align}
    \nabla_{\bm{\rho}_i} J_\text{cd} \approx \frac{J_\text{cd}(\bm{\rho}_i^+, \textsc{MP}(\bm{\rho}_i^+; \textsc{Task})) - J_\text{cd}(\bm{\rho}, \textsc{MP}(\bm{\rho}; \textsc{Task}))}{\epsilon}
\end{align}
where $\bm{\rho}_i^+$ is the design vector with $\epsilon$ added to its $i^\text{th}$ element. Using the derivative $\nabla_{\bm{\rho}} J_\text{cd}$, we can then optimize the design with gradient-based optimization.

This approach directly considers the motion as a function of the motion planner and does not assume a particular form of motion planning. Thus it allows us to use the full-body dynamics, friction cone constraints, control and state bounds in a nonlinear optimal control formulation (motion planner). This is in contrast to previous fixed-point approaches (\cite{ha_computational_2018}, \cite{desai_interactive_2018}, \cite{geilinger_skaterbots:_2018}, \cite{geilinger_computational_2020}) in which (i) the update rule needs to be derived manually for the used motion planner and (ii) arbitrary design constraints (on the vector $\bm{\rho}$) are not supported.

\subsection{Co-design: Upper Level}
We focus our work on improving the design of the $12$ Degrees of Freedom (DoFs) \textsc{Solo} robot~\cite{grimminger_open_2020}.
Particularly, we are interested in quadrupedal locomotion gaits such as trotting and jumping. To plan for these gaits, the motion planner takes as parameters the following:

\begin{itemize}
    \item The task, which is the desired gait, consisting of the contact sequence  and timings 
    \item The initial joint configuration $\bm{q}_0$
    \item The robot's joint limits 
\end{itemize}

Each of these are computed in the upper level from the design vector $\bm{\rho}$ and updated each time the optimizer calls the motion planner to compute the optimal trajectory. We compute the initial state of the robot $\bm{q}_0$ using inverse kinematics so that the angle at the knee joint of the shortest leg is $45^{\circ}$. We then run forward kinematics to set the foot positions, gait sequence and timings based on the task. We used the library \textsc{Pinocchio}~\cite{carpentier2019pinocchio} for computing the robot's kinematics and dynamics. We also set the lower and upper control bounds ($\underline{\vu}, \bar{\vu}$), and finally compute the optimal motion. We present an overview of our algorithm in~\Cref{alg:dddp}. In the upper level, we use the interior-point/direct algorithm provided in \textsc{Knitro}~\cite{byrd2006k}, which requires the derivatives of the motion planner using the parallel scheme described.

\subsection{Motion Planning: Lower Level}\label{sec:motion_planning}
The lower level of our co-design bilevel optimization algorithm computes the motion trajectory $\{\bm{X}, \bm{U}\}$ given a task and design $\bm{\rho}$.
We formulate this lower level optimization as a hybrid nonlinear optimal control problem with fixed contact sequence and timings (\Cref{eq:motion_planning_formulation}):
\allowdisplaybreaks
\begin{align}\nonumber
        \argmin_{\bm{X}, \bm{U}}&\ \sum_{k=0}^{N-1} \abs{\vq_k \ominus \vq_\text{ref}}_{\bm{Q}}^2 + \abs{\vv_k}_{\bm{N}}^2 +\abs{\vu_k}^2_{\bm{R}} + \abs{\vlam_k}_{\bm{K}} \\\nonumber
        \text{s.t.}\\\nonumber
        & \text{for each contact phase: $p\in\mathcal{P}=\{1,2,\cdots,N_p\}$}\\\nonumber
        & \hspace{1em}\text{if $\underline{\Delta}t_p\leq k\leq \bar{\Delta}t_p$:}\\\nonumber
        & \hspace{2em}\vq_{k + 1} = \vq_k \oplus \int_{t_k}^{t_k + \Delta t_k} \hspace*{-2em} \vv_k\ dt
        \tag*{(integrator)}, \\\nonumber
        & \hspace{2em}\vv_{k + 1} = \vv_k + \int_{t_k}^{t_k + \Delta t_k} \hspace*{-2em} \dot{\vv}_k\ dt, \\
        & \hspace{2em}(\vqddot_k,\vlam_k) = \bm{f}_p(\vq_k, \vqdot_k, \vu_k), \tag*{(contact dyn.)}\\\nonumber
        & \hspace{1em}\text{else:}\\\nonumber
        & \hspace{2em}\vq_{k+1} = \vq_k,\\
        & \hspace{2em}(\vqdot_{k+1},\vlam_k) = \bm{\Delta}_{p}(\vq_k, \vqdot_k), \tag*{(impulse dyn.)}\\
        & \bm{g}(\vq_k, \vqdot_k, \vu_k) = \bm{0}, \tag*{(equality)}\\
        & \bm{h}(\vq_k, \vqdot_k, \vu_k) \leq \bm{0}, \tag*{(inequality)}\\
        & \underline{\vx} \leq \vx_k \leq \bar{\vx}, \tag*{(state bounds)}\\
        & \underline{\vu} \leq \vu_k \leq \bar{\vu}. \tag*{(control bounds)}\\
        \label{eq:motion_planning_formulation}
    \end{align}
The state $(\bm{q},\bm{v})\in X$ lies in a differential manifold formed by the configuration $\bm{q}\in\mathbb{SE}(3)\times\mathbb{R}^{n_j}$ and its tangent vector $\bm{v}\in\mathbb{R}^{n_x}$ (with $n_x$ and $n_j$ as the dimension of the state manifold and number of joints, respectively).
The control $\bm{u}\in\mathbb{R}^{n_j}$ is the vector of input torques, $\vlam_k$ is the vector of contact forces,
$\ominus$ and $\oplus$ are the \textit{difference} and \textit{integration} operators of the state manifold, respectively. Then $\vq_\text{ref}$ is the reference standing upright robot posture, and $\bm{f}_p(\cdot)$ represents the contact dynamics under the phase $p$. To account for effects of discrete contact changes,  $\bm{\Delta}_{p}(\cdot)$ is used to define an autonomous system that describes the contact-gain transition (\cite{featherstone_rbdbook}).
$\bm{Q}$, $\bm{N}$, $\bm{R}$ and $\bm{K}$ are positive-define weighting matrices, $(\underline{\vx},\bar{\vx})$ and $(\underline{\vu},\bar{\vu})$ are the lower and upper bounds of the system state and control.
$\underline{\Delta}t_p$ and $\bar{\Delta}t_p$ defines the timings of the contact phase $p$. We compute the hybrid dynamics and its derivatives as described in~\cite{mastalli_crocoddyl_2020}.

During contact phases, we use a linearized friction-cone constraint via a ($\bm{A} \vlam_{\mathcal{C}(k)} \leq \bm{r}$),
where $(\bm{A}, \bm{r})$ are computed from a predefined number of edges, and minimum and maximum normal contact forces, respectively.
$\mathcal{C}(k)$ describes the set of active contacts.
During the swing phases, we also include contact-placement constraints ($\log{(\bm{p}_{\mathcal{G}(k)}^{-1} \circ \bM_{\bf{p}_{\mathcal{G}(k)}})} = \bm{0}$),
where $\log(\cdot)$ describes the log operator used in Lie algebra, $\bm{p}_{\mathcal{G}(k)}$ and $\bM_{\bf{p}_{\mathcal{G}(k)}}$ are the reference and current placements of the set of swing contacts $\mathcal{G}(k)$.

We solve the motion planning problem (\eref{eq:motion_planning_formulation}) with the Feasibility-Driven Control-limited DDP (\textsc{Box-FDDP}) algorithm~\cite{mastalli_boxfddp_2021}, a variant of the Differential Dynamic Programming (DDP) algorithm.
\textsc{Box-FDDP} uses direct-indirect hybridization and enforces hard-constraints for the control limits.
We employ a soft quadratic barrier to enforce inequality, equality and state constraints defined in~\eref{eq:motion_planning_formulation}.
We implemented the algorithm using the open-source library \textsc{Crocoddyl}~\cite{mastalli_crocoddyl_2020}.

\begin{algorithm}[t]
    \caption{Co-design optimization}
    \label{alg:dddp}
    \begin{algorithmic}[1]
    \Procedure{MP}{$\bm{\rho}$; \textsc{Task}}
      \State Compute initial state $\bm{q}_0$ using inverse kinematics
      \State Set control bounds $\underline{\vu}, \bar{\vu}$ based on actuator parameters
      \State Run forward kinematics on $\bm{q}_0$ and set foot positions
      \State Set gait sequence and timings based on $\textsc{Task}$
      \State Compute $\bm{m}$, the optimal motion
      \State \Return $\bm{m}$
    \EndProcedure
    
    \Procedure{CoDesign}{}
    \State Start at a design $\bm{\rho} = \bm{\rho}_0$
        
        \While{$J_\text{cd}(\bm{\rho}, \textsc{MP}(\bm{\rho}; \textsc{Task}))$ decreasing}
            \State Compute $\nabla_{\bm{\rho}} J_\text{cd}$ via finite differences in parallel
            \State Update $\bm{\rho}$ using one step of the NLP solver
            \State Save the resulting motion to $\bm{m}$ and cost $J_\text{cd}$
        \EndWhile

        \State \Return $(\bm{\rho}, J_\text{cd})$ -- optimal design and its cost value
    \EndProcedure
    \end{algorithmic}
\end{algorithm}

\subsection{Verification in Simulation}
We also validated our design improvements in the \textsc{PyBullet} physics simulator. (\cite{coumans_pybullet_2016}).
To do so, we execute the motion plan for both the nominal and the optimized designs, and record the percentage improvement in costs $\Delta J_\text{cd}$ (similar to~\cite{pecyna_deep_2019}).
We use a proportional-derivative (PD) controller with feed-forward torque to track the planned motion:
\begin{align*}
    \vu &= \vu^* + \bm{K_p}(\vq_j^* - \vq_j) + \bm{K_d} (\vqdot_j^* - \vqdot_j),
\end{align*}
where $\vu^*$, $\vq_j^*$ and $\vqdot^*$ are the reference feed-forward command, joint positions and velocities computed in~\eref{eq:motion_planning_formulation}, respectively. $\bm{K_p}$ and $\bm{K_d}$ are the PD gains.
We tune these gains through a grid search procedure.
We run the simulator on a $20\times20$ grid for $\bm{K_p} \in [1, 20]$ and $\bm{K_d} \in [0.1, \bm{K_d}/ 2]$.
Then, we pick the gains that lead to the smallest tracking error for both designs.
This procedure allows us to fairly compare and account for different robot dimensions and weights, as larger robots require higher gains and vice-versa.

A designer can use this second stage to validate the correctness of the dynamics model used in motion planning and the improvements in co-design cost.

\section{Co-Design Formulation -- Robot Model, Cost Function and Constraints}
\label{sec:robot_model}

Our design vector $\bm{\rho}$ consists of the lengths of the lower- and upper-leg limbs, the x-, and z-attachment points of the legs, the trunk shape: width, height and depth.  We also model the x-, and z-positions of the two electronics boxes in the base of the robot. We thus implicitly constrain a  symmetrical design along the direction of motion (the x-direction).

Next, we use an actuator model and optimize both the gear ratio and motor mass, which are the same for all motors, for simplicity. All these properties are included in the robot model to compute masses and inertias of the relevant links. For the limbs, we scale the volume linearly with the length of the leg as a simple proxy measure for structural integrity.

\subsection{Actuator Model and Cost Function}
\label{sec:modeling}

Following \cite{fadini_computational_2020} and \cite{yesilevskiy2018energy} we model the mass of the motor $m_m$ and parameterize the control limits $\underline{\bm{u}}$ and $\overline{{\bm{u}}}$ using an exponential regression based on $m_m$. We used the regression values from \cite{fadini_computational_2020}, which were fitted on datasheets from \textit{Antigravity}, \textit{Turnigy}, \textit{MultiStar} and \textit{PropDrive}:
\begin{align}
    \overline{{\bm{u}}} = -\underline{\bm{u}} = 5.48\, m_m^{0.97}.
\end{align}
Following \cite{fadini_computational_2020}, the dynamics of the system in the motion planning phase are frictionless and the actuator model is present in the co-design cost function. Given applied controls $\vu$ at the robot's joints, the total torque at the motor ($\tau_t$) is:
\begin{equation}
    \tau_t = \frac{\vu}{n} + \tau_f,
\end{equation}
where $n$ is the gear ratio and $\tau_f$ is the friction torque. The friction torque itself models the combined Coulomb and viscous friction at the transmission, which the motor needs to overcome. Thus:
\begin{equation}
    \tau_f = \tau_\mu\, \text{sign}(\omega_m) + b\,\omega_m,
\end{equation}
where $\tau_\mu$ is the Coulomb friction parameter, $b$ is the viscous friction parameter and $\omega_m$ is the motor angular speed, which is $n$ times the joint angular speed.

We then consider three power losses -- mechanical power, Joule effect from the motor winding resistance, and friction losses from the transmission:
\begin{align}
    P_\text{mech} = \tau_f \omega_m, \quad P_\text{joule} = \frac{1}{K_m} \tau_f ^ 2,\quad P_\text{fric} = \tau_f \omega_m,
\end{align}
where $K_m = 0.15 m_m^{1.39}$ is the speed-torque gradient of the motor, again computed using an exponential regression on the motor mass.

Unlike in \cite{fadini_computational_2020}, we cannot ignore the mechanical power, as the foot start and end positions are dependent on the robot body structure and the total energy is not conserved between designs (and thus not constant). We thus follow \cite{yesilevskiy2018energy} and compute the integral of the above terms ignoring power regenerative effects, summed over each of the motors:
\begin{align}
    J_\text{cd} = \int_{t_0}^{t_N} \sum_\text{motor} P_\text{elec}+ \text{max}(P_\text{fric}, 0)\ \text{dt},
\end{align}
where $P_\text{elec} = \text{max}(P_\text{mech} + P_\text{joule}, 0)$ is the positive electrical power (as defined in  \cite{yesilevskiy2018energy}). The friction power is separate, as it is due to the transmission. We integrate over the planning horizon and sum the non-negative power of each of the 12 \textsc{Solo} motors. Thus $J_\text{cd}(\cdot)$ is the integral of these power terms, corresponding to the energy used during the motion (the total work).
Finally, we note that the \textsc{Solo} robot's actuators use a custom gearbox, thus making the gear ratio independent from the motor~\cite{grimminger_open_2020}. This allows us to treat them as separate optimization targets.

\subsection{Constraints}
We then specify constraints on the design vector $\bm{\rho}$. Firstly, we add a volumetric collision constraint on the electronics boxes, the Inertial Measurement Unit (IMU) box and the motherboard (MB) box:
\begin{equation}
    \left(x_\text{mb} - z_\text{imu}\right)^2 +
        \left(x_\text{mb} - z_\text{imu}\right)^2 \leq
            \left(r_\text{mb} + r_\text{imu}\right)^2,
\end{equation}
where $x_\text{mb}, z_\text{mb}, x_\text{imu}, z_\text{imu}$ are the coordinates of the two boxes and $r_\text{mb} = 0.0361 \text{m}, r_\text{imu} = 0.0282 \text{m}$ are the radii of the smallest circumscribed sphere around them.

Finally, we specify linear constraints on the positions of the two electronics boxes and the positions of the legs so that they are within the base of the robot:
\begin{align}\nonumber
   -\frac{w_b}{2} \leq &x_\text{imu} \leq \frac{w_b}{2}, 
        -\frac{w_b}{2} \leq x_\text{mb} \leq \frac{w_b}{2},
   -\frac{d_b}{2} \leq z_\text{imu} \leq \frac{d_b}{2}, \\\nonumber
       \ -\frac{d_b}{2} \leq &z_\text{imu} \leq \frac{d_b}{2},
   -\frac{w_b}{2} \leq x_\text{fr} \leq \frac{w_b}{2}, 
        \ -\frac{w_b}{2} \leq x_\text{hr} \leq \frac{w_b}{2},\\
   &-\frac{d_b}{2} \leq z_\text{fr} \leq \frac{d_b}{2}, 
        \ \ -\frac{d_b}{2} \leq z_\text{hr} \leq \frac{d_b}{2}
\end{align}
where $w_b$ and $d_b$ are the width and depth of the base and $x_\text{fr}, z_\text{fr}$ and $x_\text{hr}, z_\text{hr}$ are the x- and z-coordinates of the front and hind shoulders. Note these inequalities constraints are defined in the upper level optimization.

\subsection{Task Description}
We are interested in optimizing the \textsc{Solo} robot design for specific tasks. As such, we fix the task description in the lower motion planning level and optimize for the most efficient robot in terms of energy.

For trotting, the high-level motion task is to take two steps forward, each of $0.05\si{\meter}$, with a fixed step height of $0.05\si{\meter}$. The step height is fixed, as the optimal step height is always $0 \si{\meter}$.
We allocated $22$ and $37$ knots\footnote{Knots are points in time for the discretization of the optimal control problem.} for the swing and double support phases of the motion, respectively, and used a symplectic Euler integrator with time-step of $10\si{\milli\second}$.

For jumping, the task is to jump forward $0.1\si{\meter}$ with a step height of $0.15\si{\meter}$.
We used the same integrator and time-step as in the trotting case.
We defined $20$ knots for the flight phase and $40$ knots for the take-off and landing phases.

Finally, for both tasks, the initial design parameters $\bm{\rho}_0$ were matched to the Solo robot design.

\subsection{Results}
\begin{figure}[t]
    \begin{subfigure}{0.485\textwidth}
        \centering
        \includegraphics[width=\textwidth]{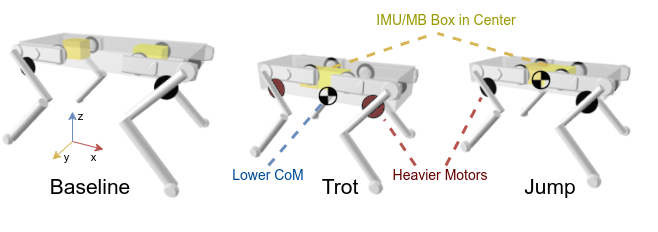}
        \caption{Resulting robot designs for trotting and jumping.}
        \label{fig:robots}
    \end{subfigure}
    \begin{subfigure}{0.485\textwidth}
        \centering
        \includegraphics[width=\textwidth,trim={0.8cm 0.0cm 0.0cm 0.0cm},clip]{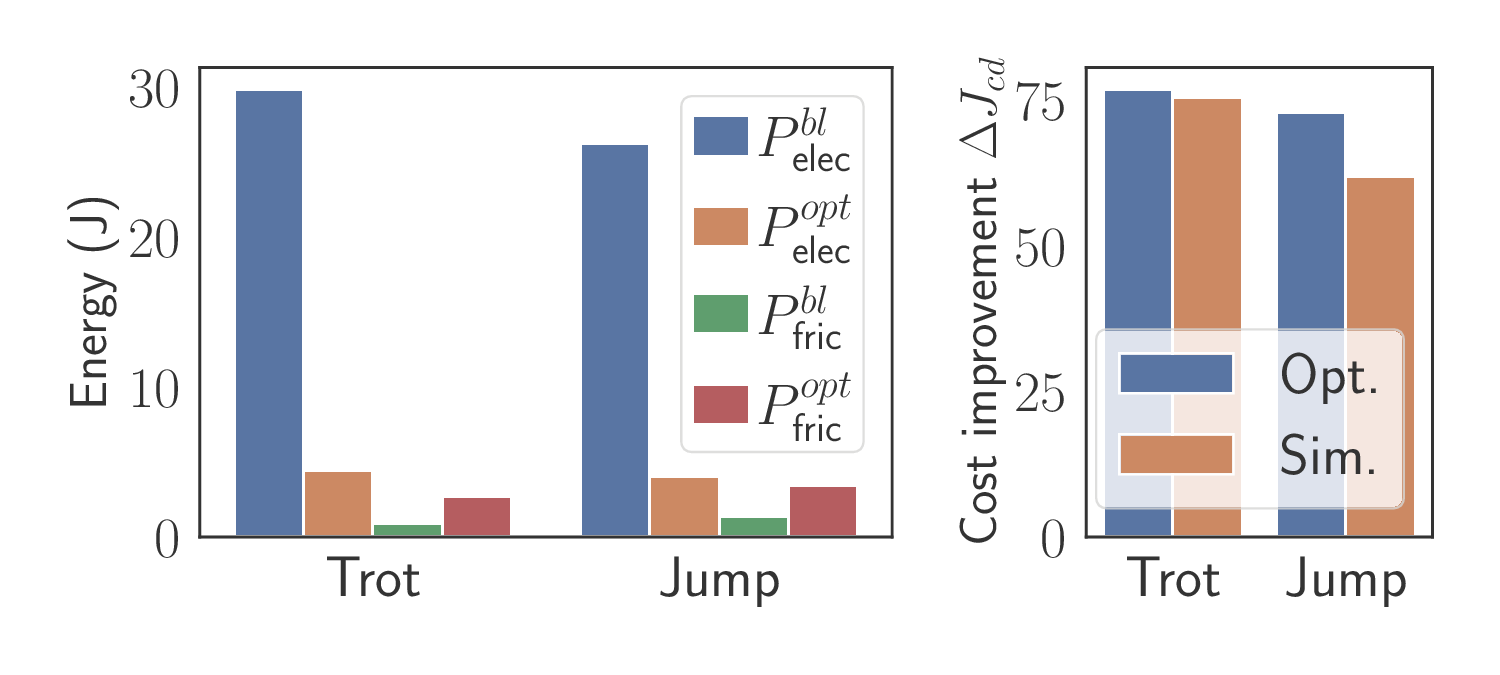}
        \caption{Cost improvements.}
        \label{fig:results}
    \end{subfigure}
    \caption{Robot designs and cost improvements on the trotting and jumping tasks. The costs are broken down for electric and friction contributions. We show the optimization and simulation percentage improvement on the bottom right.}
\end{figure}

The resulting robot designs and cost improvements are in \Cref{fig:robots,fig:results}. For both trotting and jumping, we plotted the energy contributions from the positive electrical power at the motor as $P_\text{elec}$ versus the friction contribution from the transmission as $P_\text{fric}$. The algorithm chooses to minimize the electro-mechanical losses while increasing the friction losses. This is similar to \cite{fadini_computational_2020}, as small motors are much more energy inefficient since the reciprocal of the speed-torque gradient exponentially decreases 
($K_m = 0.15 m_m^{1.39}$), increasing the Joule losses.

For trotting specifically, the friction losses are smaller, as trotting is a more static motion with smaller motor velocities, and friction is velocity-dependent. Thus the dominating cost is the electro-mechanical energy. This allows for a heavier robot with bigger motors than the optimal design for a jumping task -- the optimal motor mass is $m_m = 0.179 \text{kg}$ and gear ratio is $N = 16.062$ with a total robot weight of $3.805 \text{kg}$. The initial motor mass and gear ratio for the \textsc{Solo} robot are $m_m = 0.053 \text{kg}$ and $N = 9$ and the robot weighs $2.421 \text{kg}$. With a higher gear ratio the optimizer reduced the electro-mechanical energy further. Furthermore, we see a increase in base depth, which allows for the upper legs to be attached higher to the base of the robot. This allows for a lower center of mass, which can increase stability.

For jumping, however, a heavy robot is not optimal, as the entire mass of the robot needs to be moved. Thus the optimizer found $m_m = 0.168$ and $N = 17.325$ with a total mass of $3.592 \text{kg}$. The robot is heavier than the baseline, however the legs and the base are smaller. Compared to the optimal trotting design, the motors are lighter, but the gear ratio for both designs is similar. For both optimal designs, notably the boxes are optimally in the middle of the robot.

Finally, for both optimal designs, we also observed that the cost improvements remain in simulation within 10\% of the ones found during optimization.

\subsection{Optimality and Scalability}
\label{sec:cma}
We compared our gradient-based co-design approach to the CMA-ES genetic algorithm on the trotting task in order to check convergence properties and optimality. We used the open-source CMA-ES library \textsc{PyCMA}~\cite{niko_cma-espycma_2020}.  In order to evaluate scalability, we varied the dimensions of the co-design vector by including subsets of the decision variables, namely:

\begin{enumerate}
    \item $\dim(\bm{\rho}) = 4$ -- leg lengths (front and back)
    \item $\dim(\bm{\rho}) = 6$ -- same as $4$, and motor mass and gear ratio
    \item $\dim(\bm{\rho}) = 9$ -- same as $6$, and base shape
    \item $\dim(\bm{\rho}) = 13$ --  same as $9$, and electronics boxes
    \item $\dim(\bm{\rho}) = 17$ --  full model
\end{enumerate}

\begin{figure}[t]
    \centering
    \includegraphics[width=0.485\textwidth,trim={1cm 1.0cm 1.0cm 1.0cm},clip]{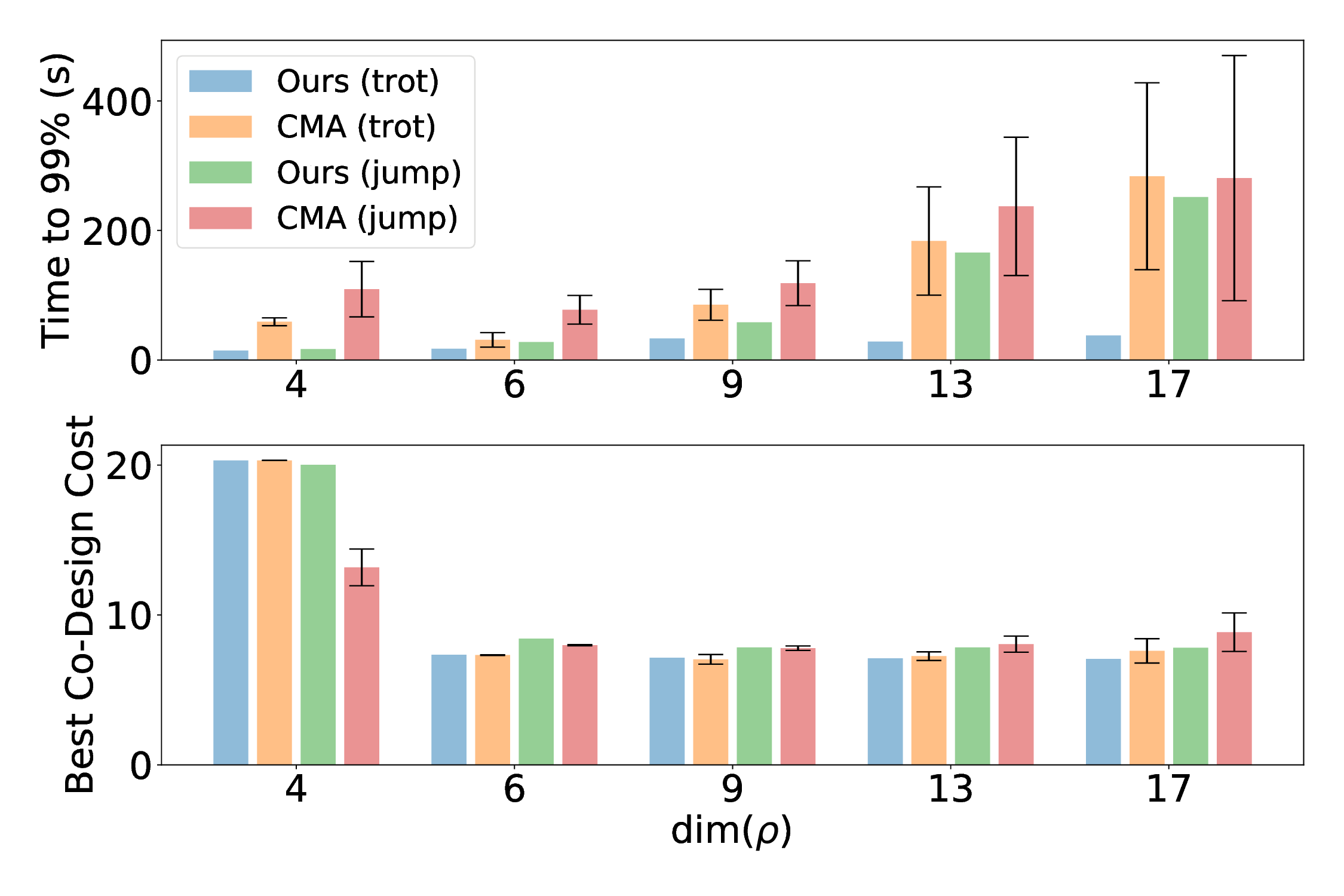}
    \caption{Scalability results for different problem dimensions.}
    \label{fig:cma_vs_gd}
\end{figure}
For CMA-ES we specified a quadratic soft penalty for all constraints. We ran CMA-ES with population sizes $N = [10, 20, 50]$ and selected $N = 50$, which achieved the same or lower costs than our approach on all problems.

Importantly, this corresponds to $50$ calls to the motion planner by CMA at each iteration versus $\dim(\bm{\rho}) + 1$ calls for our approach. Thus we measure time to convergence, as a per-iteration measure would favor our approach heavily. Both approaches used multi-threading with $8$ threads and were given the same computational budget.

We then plot the mean and standard deviation for costs and time to 99\% convergence over $20$ runs at $N = 50$ in \fref{fig:cma_vs_gd}. On the trotting task our approach has better scalability than CMA-ES, which is expected given the convergence properties of CMA-ES. On the jumping task convergence is slower for both with CMA-ES having a large deviation in convergence time for larger problems. Importantly, CMA-ES is not deterministic and although the average time for the complex jumping task is comparable to our approach, the worst-case time we observed is $600$ seconds for CMA versus $252$ seconds for our approach (both for the $17$-DOF jumping co-design task).

Finally, of interest is that we are able to achieve similar best co-design costs as CMA-ES across problem dimensions for the given co-design problems. This could indicate that our local gradient-based bilevel approach can achieve globally optimal solutions in practice, for problems like the ones studied here. %

\section{Conclusion}
In this paper we proposed a modular co-design framework for dynamic quadruped locomotion.
Our approach is based on bilevel optimization and exploits the derivatives of a hybrid nonlinear optimal control problem (lower level problem that describes the motion planner).
Our algorithm allows for the use of complex, state-of-the-art motion planners in the co-design loop together with linear and nonlinear design constraints in the upper level. One advantage of using DDP-style motion planning in our work is the guaranteed physical feasibility of the motion. When using other motion planners, this consistency might not be guaranteed and the resulting gradients might be noisy if the motion constraints are not satisfied.
We demonstrated that a coupling between the upper and lower level costs is beneficial.
Note that we have a weak coupling, where the lower level has a regularization on the square of the torques and the upper level has the Joule effect cost, also on the square of the torques.

Future work lies in using analytical derivatives instead of using finite differences, which introduce numerical errors when computing the derivative of the motion planner. Furthermore, our approach can use any differentiable motion planner and there are interesting opportunities in using different motion planners and formulations (for instance different contact models and constraints) which can enable co-design in more complex domains, for instance with sliding or slipping contacts. Additionally, of interest is handling more complex state constraints that come from the environment, for instance for footstep planning -- determining the contact locations and timings of footsteps. %

\bibliographystyle{IEEEtran}
\bibliography{IEEEfull,IEEEconf,codesign,manual}

\end{document}